# Analyze the robustness of three NMF algorithms (Robust NMF with L1 norm, L2-1 norm NMF, L2 NMF)

Author Name：Cheng Zeng, Jiaqi Tian, Yixuan Xu

Email: czen8507@uni.sydney.edu.au , jtia3555@uni.sydney.edu.au, yixu9725@uni.sydney.edu.au

This paper consists of 4560 words (excluding Appendix) and 19 pages (excluding Appendix).

## 1, Abstract

Non-negative matrix factorization (NMF) and its variants have been widely employed in clustering and classification tasks (Long, & Jian , 2021). However, noises can seriously affect the results of our experiments. Our research is dedicated to investigating the noise robustness of non-negative matrix factorization (NMF) in the face of different types of noise. Specifically, we adopt three different NMF algorithms, namely L1 NMF, L2 NMF, and L21 NMF, and use the ORL and YaleB data sets to simulate a series of experiments with salt-and-pepper noise and Block-occlusion noise separately. In the experiment, we use a variety of evaluation indicators, including root mean square error (RMSE), accuracy (ACC), and normalized mutual information (NMI), to evaluate the performance of different NMF algorithms in noisy environments. Through these indicators, we quantify the resistance of NMF algorithms to noise and gain insights into their feasibility in practical applications.

## 2, Introduction

Dictionary Learning represents data as a linear combination of a set of basis vectors (dictionaries). Within the framework of dictionary learning, sparse coding methods are usually used, such as sparse coding or K-SVD, to acquire knowledge about both the dictionary and the associated sparse coefficients (Wu, D., Zhao, P., & Wan, Q. (2023)). On the other hand, Non-negative Matrix Factorization (NMF) represents a specific matrix factorization approach, wherein a non-negative data matrix is approximated as the product of two non-negative, low-dimensional matrices. In the context of NMF, these two matrices correspond respectively to the "basis" and "coefficients" of the data, which is similar to basis vectors and coefficients in dictionary learning. Notably, NMF enjoys widespread applicability across various domains and is particularly well-suited for handling non-negative data, as well as for performing data dimensionality reduction and feature extraction.

Non-negative Matrix Factorization (NMF) represents data as a linear combination of a set of non-negative basis vectors. The fundamental concept behind NMF revolves around approximating the original data by identifying a set of non-negative basis vectors and coefficients, aiming to provide a more insightful understanding of the intrinsic structure within the data (Greene and Cunningham, 2009). NMF finds extensive application in various domains such as image processing, text mining, bioinformatics, and more, where it is employed for tasks including feature extraction, topic modeling, data compression, and others.

However, a key issue related to NMF is the performance issue under noisy data conditions. Given NMF's assumption of non-negativity in data, the presence of noise in the data can potentially lead to inaccuracies and stability issues during the factorization process. Noise significantly impacts the results of NMF, resulting in inaccurate representations, particularly for applications requiring high precision. For instance, salt-and-pepper noise may degrade image quality during image processing, and NMF may struggle to handle such scenarios effectively, leading to artifacts and distortions. This challenge restricts the reliability of NMF in practical applications. Therefore, noise modeling and robustness enhancement remain focal points of research to enhance the performance of NMF in noisy environments.

Our aim is to address the issue of robustness in NMF under noisy environments. Noise adversely affects the performance of NMF and makes it more difficult to extract useful information from noisy data. To ensure the reliability of the NMF algorithm in practical applications, we reproduce the classic NMF algorithm and conduct a series of experiments, utilizing various NMF algorithms to analyze the impact of different types of noise and datasets on NMF performance. Therefore, noise modeling and denoising techniques are used in NMF to improve the robustness and effectiveness of NMF in the face of noisy data.

Finally, 16 experiments were conducted by setting different parameters, and the RMSE, ACC, and NMI values of the three algorithms were compared. This result suggests that for us to further investigate and explore our research questions in this experiment, L2 NMF would be the most suitable choice among the three models when using large dataset, but L21is the best when using smaller dataset. Its superior performance demonstrates its potential to effectively handle analysis problems with different data sets in different types of noises.

## 3, Related Work

NMF (non-negative matrix factorization) is a popular unsupervised learning algorithm for high dimensional data because it is able to effectively factorize a non-negative matrix into two lower dimensional non-negative matrices. Due to its outstanding ability to handle sparsity and select significant features, NMF is widely used in a variety of applications, such as image preprocessing and bioinformatics.

The NMF is:

$$X \approx RD \qquad (1)$$

Where X is a non-negative matrix( it represents data, and in our project, it is the image), NMF aims to find two optimal non-negative matrices R and D, whose product is as close as possible to X. R is usually features of X, and D is commonly used as a weight matrix to value features.

The objective function is :

$$min \parallel X - RD \parallel_F^2 \qquad (2)$$

Our goal is to minimize the Frobenius norm of matrix difference between the original matrix and the reconstructed matrix.

In our project, we mainly the utilize multiplicative update rule to optimize our NMF methods. MUR uses multiplicative factors to update R and D iteratively in regard to the minimization of the objective function(ScienceDirect Topic, 2013). The MUR is like:

$$R = R_{ij} \frac{(XD^T)_{ij}}{(RDD^T)_{ij}} \qquad (3)$$

$$D = D_{ij} \frac{(R^T X)_{ij}}{(R^T RD)_{ij}} \qquad (4)$$

This is what R and D look like in L2 norm NMF using MUR. The main idea behind MUR is to consistently hold, but some modifications are needed when applying MUR to different NMF algorithms. For instance, in the L2,1 robust NMF algorithm, a diagonal matrix is essential to collaborate with R and D on the optimization of the objective function.

# 4, Methods

## 4.1 Preprocessing

Preprocessing is a very important step before doing a machine learning algorithm. During the preprocessing step, Many decisions that can affect the prediction results of the model are made (Gonzalez Zelaya, 2019).

Data preprocessing methods include data cleaning, normalization, standardization, dimensionality reduction, data partitioning, data centering, and data enhancement (García, Julián Luengo and Herrera, 2015). They are usually used to eliminate unnecessary noise, reduce computational complexity, improve model stability, etc.

In this experiment, we chose to use normalization to preprocess the data. It can scale image data to the range of [0, 1]. Therefore, it can reduce the scale difference between different features, thereby improving the performance of the model. Because the algorithm we need to explore is NMF, the data is required to be non-negative. Normalization can meet this requirement. We also try to use Data Centering to preprocess the data. However, it was found that the result was not as good as Normalization. Because centering requires subtracting the mean of the data, this may result in a negative number, thus violating the non-negativity of NMF.

Use Python code 'X_noise = np.random.rand(*X_hat.shape) * 40' to create noise and add it to the ORL data set. The following data can be used to compare the RMSE performance of different preprocessing:

1. No processing, rmse: 37.2

2. Normalization: 19.9

3. normalization + centering: 20.5

4. Data centering: 25

## 4.2 L2 NMF (Standard NMF)

The standard NMF approach and its variants have been extensively used as feature extraction techniques for various applications, especially for high dimensional data analysis. The newly formed low-dimensionality subspace represented by the basic vectors should capture the essential structure of the input data as best as possible. (Buciu, 2008) Its original design intention is to achieve data representation and dimensionality reduction by finding basic components with non-negativity constraints in data analysis and signal processing, to better understand and interpret the original data. The conscious implication of the design principle is to ensure practical interpretability of decomposition results by maintaining non-negativity constraints and is often used in application areas that deal with non-negative data.

### 4.2.1 Objective function

The loss function commonly used by L2 Non-negative Matrix Factorization (Standard NMF) can be expressed as the following mathematical expression. Given a non-negative matrix $V \in \mathbb{R}^{m \times n}$, and the non-negative matrices $W \in \mathbb{R}^{m \times r}$ and $H \in \mathbb{R}^{r \times n}$ to be decomposed into, The loss function can be defined as:

$$L(V, W, H) = \frac{1}{2} \parallel V - WH \parallel_F^2 \tag{5}$$

Among them, $\| \cdot \|_F$ represents the Frobenius norm, which is the F norm of the matrix, which measures the difference between the two matrices. The goal of this loss function is to minimize the difference between the original matrix $V$ and the decomposed matrix $WH$, by adjusting the values of $W$ and $H$.

### 4.2.2 Optimization

In L2 Non-negative Matrix Factorization (Standard NMF), in order to minimize the above loss function, the gradient descent method or other optimization algorithms are usually used to update the values of the matrices $W$ and $H$.

For the update rules of matrix W:

$$W_{ij} \leftarrow W_{ij} \cdot \frac{(VH^T)_{ij}}{(WHH^T)_{ij}} \tag{6}$$

For the update rules of matrix H:

$$H_{ij} \leftarrow H_{ij} \cdot \frac{(W^T V)_{ij}}{(W^T WH)_{ij}} \tag{7}$$

These rules update the values of W and H through repeated iterations until convergence conditions are reached or a predetermined number of iterations is reached.

### 4.2.3 Advantage

1. STANDARD NMF decomposition is interpretable because the obtained decomposition matrices W and H are both non-negative and can be used for interpretation and visualization of data features.

2. In many applications, STANDARD NMF is able to extract latent features and reduce data dimensions, thereby helping to reduce computing and storage costs.

### 4.2.4 Disadvantage

1. The results of STANDARD NMF may be affected by the choice of initial values and the number of iterations, and parameters need to be carefully selected to obtain the best results.

2. STANDARD NMF tends to generate dense decomposition matrices, which may result in large computational and storage overhead.

3. For some data sets, STANDARD NMF may not be flexible enough to capture complex data structures, and other decomposition methods may need to be tried.

### 4.3 L2-1 NMF

### 4.3.1 Objective function

When we investigate the results of L2 norm NMF, we find that the L2 norm NMF is prone to outliers and noise. The error function of L2 norm NMF is:

$$\|X - RD\|_F^2 = \sum_{i=1}^{n} \| x_i - Rd_i \|^2 \tag{8}$$

From the function we know that the residual error is squared for each data point, leading to a result that a few large outliers can have a huge impact on the performance of NMF. Therefore, we introduce a robust NMF-L2,1 to mitigate the influence of outliers and improve the performance of NMF against different noises. The error function of L2,1 robust NMF is:

$$\left\| X - RD \right\|_{2,1} = \sum_{i=1}^{n} \sqrt{\sum_{j=1}^{p} (X - RD)_{ji}^2} = \sum_{i=1}^{n} \| x_i - Rd_i \| \tag{9}$$

As we can see, the error for each data point is not squared, therefore, L2,1 robust NMF is more resilient to outliers and noises than the standard NMF.

The L2,1 robust NMF is formulated as:

$$min_{R,D} \|X - RD\|_{2,1} \quad s.t. \ R \geq 0, D \geq 0 \tag{10}$$

Our goal is to find a way to minimize the difference between X and RD. In standard NMF, the multiplicative update rule is derived because L2 norm is smooth everywhere. However, MUR is not appropriate for L2,1 because L2,1 norm combines L2 norm and L1 norm, which is not smooth and differentiable everywhere, making direct application of MUR difficult. Thus, we introduce matrix H to design a new update rule for our loss function. The matrix H is defined as:

$$H_{ii} = \frac{1}{\sqrt{\sum_{j=1}^{p}(X-RD)_{ji}^2}} = \frac{1}{\|x_i - Rd_i\|} \tag{11}$$

H is a diagonal matrix, and it can be used in L2 norm to reduce the effect of outliers. According to the work of Deguang Kong (2011), R and D are derived as:

$$R_{jk} \Leftarrow R_{jk} \frac{(XHD^T)_{jk}}{(RDHD^T)_{jk}} \tag{12}$$

$$D_{ki} \Leftarrow D_{ki} \frac{(R^T XH)_{ki}}{(R^T RDH)_{ki}} \tag{13}$$

one of the most significant advantages of L2,1 robust NMF is that it is not only resilient to outliers, but also it is adaptable and can be applied in many fields with almost the same computational cost as L1 norm NMF.

### 4.3.2 Optimization

Firstly, we calculate the diagonal matrix H, which provides weights to regularize outliers within features. Then we update D while fixing R and update R while fixing D. The correctness of the algorithm is proved by Karush-Kohn-Tucker condition of the constrained optimization theory.

### 4.3.3 Advantages

Unlike L2 norm NMF, L2,1 robust NMF utilizes a diagonal matrix H as weights to suppress outliers (Deguang Kong, 2011). It can be effectively applied in a variety of fields such as feature selection and image processing because of its accuracy and low computational cost. In addition, L2,1 robust NMF combines benefits of both L1 norm and L2 norm, so it is especially reliable in significant feature selection and preventing overfitting.

## 4.4 Robust NMF with L1 Regularization

Robust NMF with L1 norm is an algorithm that can often improve the robustness of NMF without knowing the location of the noise. In real life, data samples are prone to be partially damaged, such as salt and pepper noise, block occlusion noise, etc (Shen et al., 2014). To make matters worse, often the location of the noise is unknown, which makes NMF perform poorly when processing this data. Therefore, the robust NMF with L1 norm algorithm can better handle partially corrupted data by adding matrix E, $E \in \mathbb{R}^{m*n}$ ,which is a large additive noise (Shen et al., 2014).

### 4.4.1 Objective Function

In NMF, it is common to split a given nonnegative matrix $X$ into the product of two nonnegative proofs $U$ and $V$ (Lee and Seung, 2000)

$$X = UV \tag{14}$$

where: $X \in \mathbb{R}^{m*n}$, $U \in \mathbb{R}^{m*k}$, and $V \in \mathbb{R}^{k*n}$

In the Robust NMF with L1 Norm algorithm, it is assumed that $\hat{X}$ is a clean matrix not contaminated by noise, and $E$ is a large additive noise. Therefore, we have $X = \hat{X} + E$.

Like traditional NMF, $U$ dot multiplied by $V$ can be approximated to clean data $\hat{X}$, therefore we have:

$$X \approx UV + E \tag{15}$$

The purpose of the objective function is to guide the algorithm to continuously optimize in the ideal direction. The objective function of robust NMF and L1 norm is written as:

$$O_{RobustNMFL1} = ||X - UV - E||_F^2 + \lambda \sum_j [||E_{\cdot j}||_0]^2 \qquad (16)$$

It mainly contains two components: loss function and regularization term. The main purpose of the first part is to minimize reconstruction errors (Shen et al., 2014). Make sure $UV + E$ is as close as possible to the observed data $X$

The purpose of the second part is to impose a penalty on the complexity of the model, specifically on the sparsity of the noise matrix $E$(Shen et al., 2014). $\lambda$ is used to control the weight of the regularization term in the objective function.

However, the L0 norm is very tricky in optimization because it causes a non-convex problem. Therefore, according to Shen et al. (2014), L1 norm is used to approximate L0 norm. Let $E = E_p - E_n$, $E_p = \frac{|E|+E}{2}$, $E_n = \frac{|E|-E}{2}$, and $E_p \geq 0, E_n \geq 0$. Therefore, we have optimization function like:

$$\begin{aligned} O_{RobustNMFL1} &= ||X - UV - E||_F^2 + \lambda \sum_j [||E_{\cdot j}||_1]^2 \\ &= ||X - [U, I, -I]\begin{pmatrix} V \\ E_p \\ E_n \end{pmatrix}||_F^2 + \lambda \sum_j [||E^p_{\cdot j}||_1 + ||E^n_{\cdot j}||_1]^2 \end{aligned} \qquad (17)$$

Because $E$ can be negative or non-negative. Therefore, $E$ is decomposed into two non-negative matrices $E_p$ and $E_n$ to facilitate optimization. In addition, we need to ensure that $\hat{X}$ is a non-negative matrix, and $\hat{X} = X - E$ needs to be greater than 0.

In short, the objective function should be minimized under the constraints of ensuring that $U, V, E_p$, and $E_n$ are non-negative (i.e., $U \geq 0, V \geq 0, E_p \geq 0, E_n \geq 0$) and that $X - E$ is also non-negative.

### 4.4.2 Optimization

In Equation (17), we use L1 instead of L0 to simplify our optimization. However, we still must face the problem that $O_{RobustNMFL1}$ is not convex relative to $U, V, E_p$ and $E_n$. According to Shen et al. (2014), we can use the multiplicative update algorithm to iteratively update $U, V, E_p$ and $E_n$ to find the local minimum.

**Update U**

By fixing $E_p$ and $E_n$, we can update the $U$ value.

$$O_{RobustNMFL1} = \underset{U \geq 0}{argmin}||X - [U, I, -I](\begin{pmatrix}V\\E_p\\E_n\end{pmatrix})||_F^2$$

$$+\lambda \sum_j [||E^p_{\cdot j}||_1 + ||E^n_{\cdot j}||_1]^2$$

$$= \underset{U \geq 0}{argmin}||[X - E] - UV||_F^2 \tag{18}$$

According to the derivation of Shen et al. (2014), we can get the following formula for updating $U$:

$$U_{ij} = U_{ij}\frac{(\hat{X}V^T)_{ij}}{(UVV^T)_{ij}} \tag{19}$$

Where $\hat{X} = X - E \geq 0$

**Updating V, Ep and En**

After updating $U$, we can update $V$, $E_p$, and $E_n$ at the same time. Firstly, let $\tilde{V} = \begin{pmatrix}V\\E_p\\E_n\end{pmatrix}$, $\tilde{X} = \begin{pmatrix}X\\0_{1*n}\end{pmatrix}$, $\tilde{U} = \begin{pmatrix}U, I, -I\\0_{1*k} \sqrt{\lambda}e_{1*m} \sqrt{\lambda}e_{1*m}\end{pmatrix}$, and $S_{ij} = |(\tilde{U}^T\tilde{U})_{ij}|$, then we have:

$$\tilde{V}_{ij} = max\left(0, \tilde{V}_{ij} - \frac{\tilde{V}_{ij}(\tilde{U}^T\tilde{U}\tilde{V})_{ij}}{(S\tilde{V})_{ij}} + \frac{\tilde{V}_{ij}(\tilde{U}^T\tilde{X})_{ij}}{(S\tilde{V})_{ij}}\right) \tag{20}$$

### 4.4.3 Advantage

The main advantage of the Robust NMF with L1 Norm algorithm is reflected in the processing of partially corrupted data (Shen et al., 2014). In real environments, much data may be corrupted by noise at unknown locations. Robust NMF provides us with a solution to deal with noise in parts of the image. It does not require information about the location of the noise. This method can treat partial damage as large additive noise while estimating the location and value of the noise by computing the product of two non-negative matrices U and V. This method enhances the robustness of NMF.

## 4.5 Noise

Noise can lead to loss of image information and degradation of image quality (Azzeh, Zahran and Alqadi, 2018). In daily life, images containing noise are very common. Therefore, we can evaluate how robust an algorithm is by seeing if it can still make accurate predictions when there's noise in the image.

In this experiment, we added two types of noise, block-occlusion noise and salt and pepper noise.

### 4.5.1 Block-occlusion noise

Block-occlusion noise covers a part of the image with wrong information. It can simulate the loss or damage of image information. Algorithms that can handle block-occlusion noise are often used to restore damaged images in real life, such as when a person's face is occluded by some objects (Cotter, 2010). In a two-dimensional image, if the pixel values of one or more matrix areas are lost or damaged, it is block-occlusion noise.

Its mathematical expression is as follows:

Let matrix X be the original image with size m*n, and define the block-occlusion noise matrix N to be of size m*n. Then let a x*y matrix of N be a noise area, where x<=m, y <= n. The value of this area is not 0, and the value of the remaining areas is 0. Let this noise matrix be fused with the original matrix. The areas where the noise matrix overlaps with the original matrix will be affected, while the pixel values in other areas will remain unchanged. therefore:

$$X_{noise} = X + N \qquad (21)$$

The image below shows the comparison between the image with block-occlusion noise added and the original image. The size of the occlusion area is 10*10, and the value of the occluded area is set to 0.5:

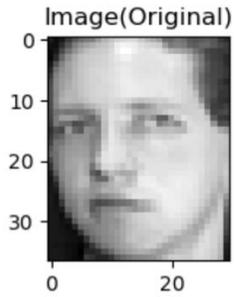 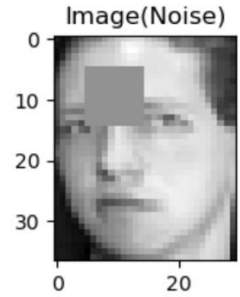

### 4.5.2 Salt and pepper noise

Salt and pepper noise is also a common type of image noise. It randomly distributes black and white pixels across the image, so it looks like salt and pepper are sprinkled across the image (Azzeh, Zahran and Alqadi, 2018). In real life, this kind of noise usually appears during image transmission. Algorithms that can handle salt and pepper noise can often be used for image enhancement.

Its mathematical expression is as follows:

Replace random pixels in the image with the maximum value 255 or the minimum value 0. Let X be the original image with dimension m*n, and define a salt and pepper noise matrix N with size m*n. A certain proportion of pixels in matrix N will be set to 255 or 0. The matrix N is then overlapped with the original matrix X. Pixels that overlap 255 or 0 in the X and N matrices will be set to 255 or 0, and the remaining pixels will remain unchanged.

$$X_{noise} = X + N \tag{22}$$

The picture below shows the comparison between the image with salt and pepper noise added and the original image. We set 40% of the pixels in the entire image to be noise, and the ratio of salt to pepper is 0.45:0.55.

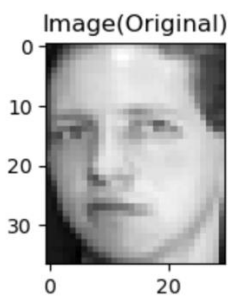 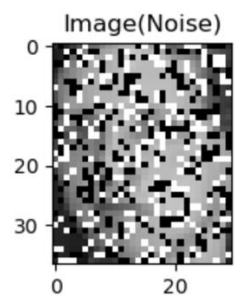

## 5, Experiment

In this project, we conduct a series of experiments to compare the performance of our three different NMF algorithms (L1 robust regularized NMF, L2 norm NMF, and L2,1 robust NMF) and explore their robustness against two types of noises (Block-Occlusion noise and Salt&Pepper noise)

The datasets used in our dataset are ORL and Extended YaleB. The ORL dataset contains 40 distinct subjects and each subject has 10 different images, which vary in the lighting, facial expression (open or closed eyes, smiling or not smiling), and facial details (glasses or no glasses). We resize the original images from 92*112 pixels to 30*37 pixels to reduce comlexity. The Extended YaleB dataset consists of 2414 images of 38 human subjects under 64 illumination conditions and 9 poses. We also resize all images to 42*48 pixels.

### 5.1, Evaluation Metrics

We implement three evaluation metrics to assess the performance of our NMF algorithms and their robustness in the presence of different noises.

1)**Root Mean Square Error (RMSE):** RMSE is commonly used to describe the difference between actual value and predicted value. It is formulated as :

$$RMSE = \sqrt{\frac{1}{n}\sum_{i=1}^{n}(actual_i - predicted_i)^2} \tag{23}$$

Based on the formula, it can be interpreted as the square root of the average of the squared residuals. Therefore, We expect a smaller RMSE value because it indicates the residuals are relatively small and the model fits well.

2)**Average Accuracy:** The average accuracy is formulated as :

$$Acc(Y, Y_{pred}) = \frac{1}{n}\sum_{i=1}^{n} 1\{Y_{pred(i)} == Y(i)\} \tag{24}$$

The datasets contain various subjects, and the average accuracy describes the proportion of subjects in our reconstructed matrix that match the subjects in the dataset through K-means clustering.

3) **Normalized Mutual Information (NMI):**

$$NMI(Y, Ypred) = \frac{2 * I(Y, Ypred)}{H(Y) + H(Ypred)} \quad (25)$$

I(.,.) is mutual information and H (.) is the entropy. NMI is commonly used to measure the similarity between two clusters.

## 5.2, Result

To optimize the performance evaluation, we randomly select 90% data as training set for the experiment and repeat the process 5 times to obtain the average evaluation metrics on different subsets. We also test two different datasets and two distinct noises to compare the robustness of three NMF algorithms. We also visualize the trend of three evaluation metrics (RMSE, Acc, NMI) in a grid of different number of components to determine the optimal choice of K(the number of components)

The number of components are 10,20,30, and 40. Maximum number of iteration is 200. For L1 NMF, the lambda is 0.1.

The line plots for ORL with block occlusion noise:

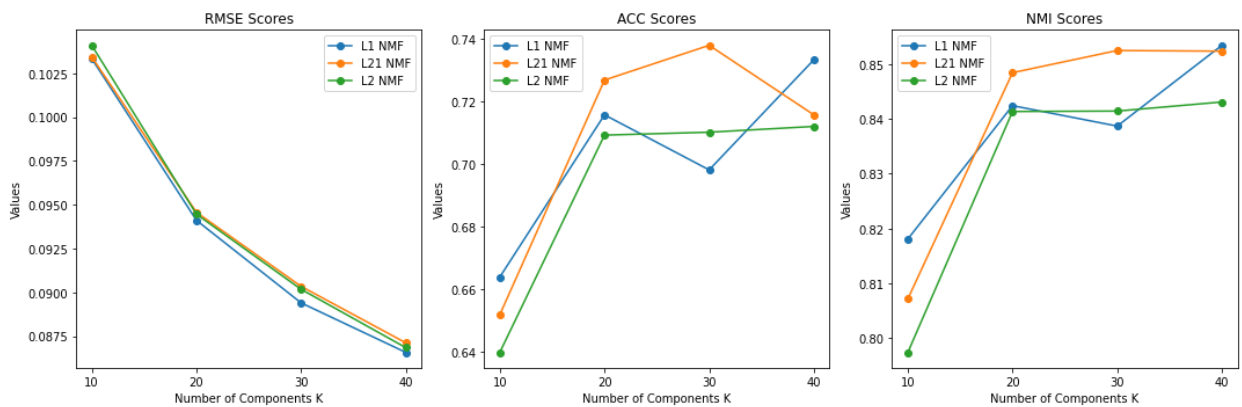

From the plots we know that all RMSE values of three NMF methods are declining as k increases. The average accuracy scores have trends of increasement as the number of components increases, however acc score of L21 reaches its highest point at k =30. NMI scores have similar trends as ACC scores.

The line plots for YaleB with block occlusion noise:

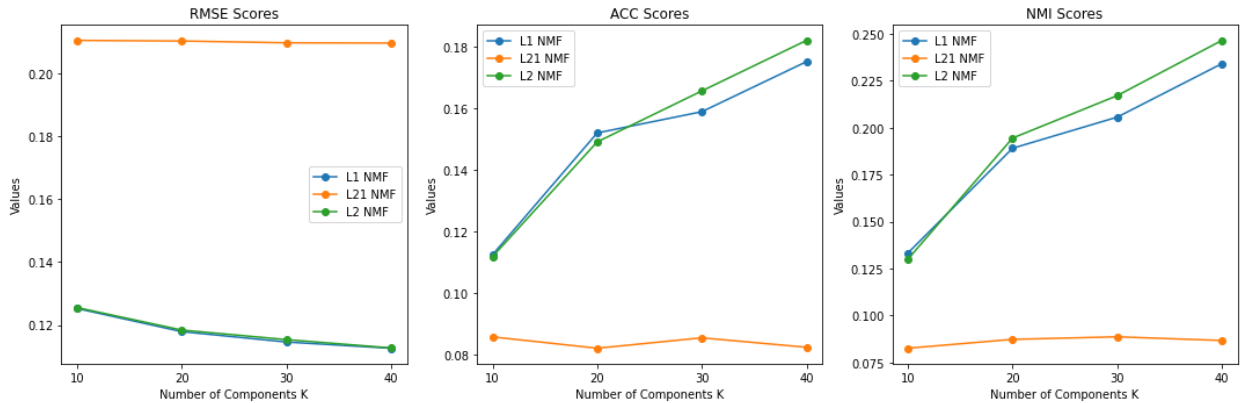

From the plot, we observe that evaluation metrics of L2,1 robust NMF barely change and have the worst performance. It is also noticeable that evaluation metrics of L1 and L2 are almost overlapped, suggesting that two methods have comparable robustness against block occlusion noise on YaleB dataset.

The line plots for ORL with Salt and Pepper noise:

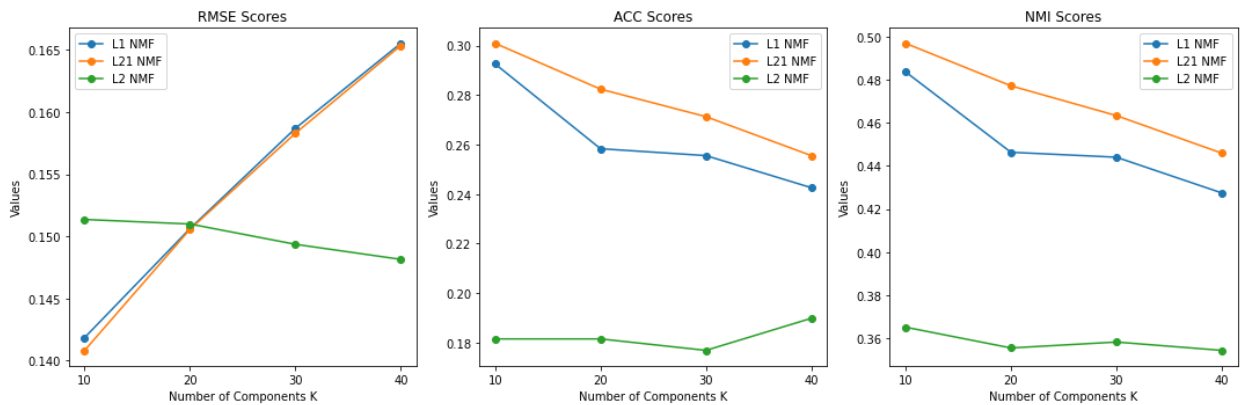

The RMSE scores plot suggests that L21 and L1 are almost indistinguishable and are increasing proportionally as K increases. But RMSE score of L2 is decreasing slowly as k becomes larger. The other two plots give solid evidence that L2 is far less effective and liable regarding salt and pepper noise than L21, which has the best performance, and L1, which is also not bad.

The line plots for YaleB with Salt and Pepper noise:

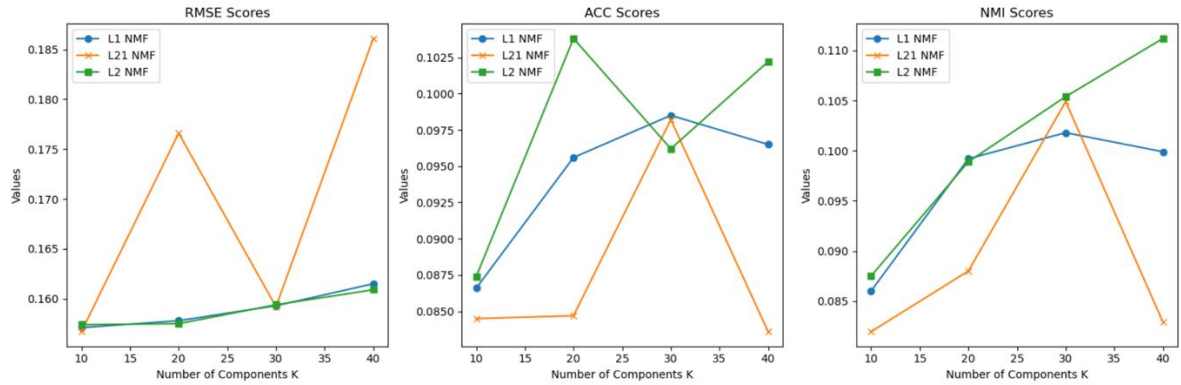

According to the first plot, we can infer that choice of k is significant since the values change dramatically and unpredictably, therefore, we need to test more choice of K to determine the optimal number of components in future experiments. The ACC and NMI plots are similar, both showing that L1 is much better than other two methods when k=10 and 20, but it suddenly drops to the same level of other two methods when K becomes bigger.

We also calculated standard deviation for evaluation metrics values. Below is an example of standard deviation for ORL with block noise:

# Standard Deviation Table

| Metrics / Methods | K=10 | K=20 | K=30 | K=40 |
| --- | --- | --- | --- | --- |
| RMSE | L1: 0.00033<br>L2: 0.00036<br>L21: 0.00024 | L1: 0.00029<br>L2: 0.00014<br>L21: 0.00032 | L1: 0.00008<br>L2: 0.00015<br>L21: 0.00030 | L1: 0.00016<br>L2: 0.00020<br>L21: 0.00017 |
| ACC | L1: 0.01636<br>L2: 0.00729<br>L21: 0.00859 | L1: 0.00797<br>L2: 0.01510<br>L21: 0.02916 | L1: 0.02058<br>L2: 0.01702<br>L21: 0.02238 | L1: 0.00989<br>L2: 0.01609<br>L21: 0.02058 |
| NMI | L1: 0.01238<br>L2: 0.01210<br>L21: 0.00601 | L1: 0.00598<br>L2: 0.01664<br>L21: 0.02017 | L1: 0.00825<br>L2: 0.00642<br>L21: 0.00980 | L1: 0.00671<br>L2: 0.01154<br>L21: 0.02043 |

All values are very small indicating that our NMF algorithms are stable.

In conclusion, L1 robust NMF is effective and generalizable to different datasets and various noises. In the future analysis, we are looking forward to leveraging L1robust NMF through a series of experiments.

## 6, Conclusion

In summary, in this project, we successfully simulated three algorithms of the NMF model to explore the impact of different types of noise on the robustness of the algorithm. We use the noise data sets ORL and YaleB. The noise types are salt-and-pepper noise and block occlusion noise. We set three evaluation indicators: root mean square error (RMSE), accuracy (ACC) and normalized mutual information (NMI). Above all, a series of simulation experiments were conducted to ensure that different noise levels and types were generated. Conduct a thorough analysis of the three algorithms L1, L2, and L21 of the NMF model. For the overall improvement of all algorithms, we note that L2 performs best when using the large YaleB dataset. Otherwise, when using smaller dataset ORL, L21 is the best.

Our experiment used two types of noise, set three evaluation indicators, and conducted 16 experiments by setting different iteration times and K values to obtain the current experimental results. Therefore, our experimental results are not just the evaluation of a single indicator. Based on the data obtained from the experiments, we can see that noise optimization has a significant impact on the robustness of the NMF algorithm.

The three NMF algorithms used this time still have a lot of room for optimization. First, we can look for more data sets with noisy functions to conduct multiple experiments to generate additional training samples, which can increase the diversity and size of the training set. Secondly, we can further in-depth parameter adjustment and method debugging to obtain more experimental samples to evaluate various NMF algorithms. At the same time, using more types of noise to train the data set to see the impact of noise on the robustness of different algorithms of the NMF model.

Third, in the future, we can extend this research in several aspects in other related directions. For example, whether the robustness of the L1NMF algorithm is affected by the regularization parameters and what the impact is. In addition, the NNDSV initialization method is extended to other NMF algorithms. (Díaz & Steele, 2021) In this experiment, we drew conclusions based on the research on the NMF algorithms. Although the applicability is not particularly high, it also provides some experimental results for subsequent related research. Reflecting on this experiment, we should further explore the opportunities for the sensitivity of the robust NMF model to the initial state of the factor matrix (Díaz & Steele, 2021), and in what fields it will be more appropriate and widely used in the future.

## 7, Reference

# 8, Appendix

## 8.1 Code introduced.

The code in this notebook is used for three NMF algorithms. And it shows how they behave on two data sets and two noises.

What I'm going to do in this section is describe the directory of this code, and how to run the code

Content page of code in ipynb:

1, Load Dataset

1.1, Load ORL Dataset and Extended YaleB Dataset

1.2 Display data

2, Method

2.1, Noise

2.2, Three NMF algorithm

3, Experiment

3.1, Training Model

3.2 Reconstruction Matrix

3.3 Visualization

4, Evaluation Metrics

4.1, Root Means Square Errors

4.2, Evaluate Clustering Performance

4.3, Evaluate RMSE, ACC, NMI for five Times (rigorous performance evaluation)

5, Appendix

**Instruction**

For data loading, we only support loading ORL and YaleB datasets. Given the structure of the folder: the code is in the algorithm folder, the data set will be placed in the data folder. So the path we read is '../data/ORL' or '../data/CroppedYaleB'

The following section describes what each section does so that you can better run the code according to your needs. You can also simply use run all to run all blocks of code in order.

Our code is made up of multiple functions, so to run the code without errors, we first need to run the first code block in 1.1 section, which is used to import all the libraries we need to use. Note that we are using the sklearn library here, but it is only used for evaluation.

Then you need to run all the functions to make sure that we don't make mistakes when we run the main program.

Functions exist with the 1.1 section, 2.1 section, 2.2 section, 3.1 section, and first code block of 4.2 section.

After that, if you want to reconstruct the image using the three NMF algorithms, you can run the code in 3.2. This part of the code includes a number of parameters:

**dataset**: is used to set the dataset you want to use, with two options: ORL and YalB

**reduce**: is used to reduce the computation complexity of images, and we recommend 3 for the ORL dataset and 4 for the YaleB dataset

**noise**: is used to set the noise you want to choose, you can choose to pass in "block_occlusion", "salt_and_pepper" or "no_noise"

**k:** is the number of component. In our report, we chose 10, 20, 30, and 40 as observation objects. You can also modify the k value to observe the prediction results

Then, after you have the reconstructed matrix X_pred, you can visualize it using code from 3.3, and calculate RMSE, ACC, and NMI using code from 4.1 and 4.2

Or you can choose to run code from 4.3 to directly test RMSE, ACC, and NMI of the three algorithms. Unlike 4.1 and 4.2, this part of the code will loop the experiment 5 times and calculate their mean and standard deviation.

There are still some parameters that you can adjust in 4.3 section:

**dataset**: is used to set the dataset you want to use, with two options: ORL and YalB

**reduce**: is used to reduce the computation complexity of images, and we recommend 3 for the ORL dataset and 4 for the YaleB dataset

**image_size**: we recommend (37, 30) for the ORL dataset and (48, 42) for the YaleB dataste

**n_iter**: number of iteration. In our report, we choose 200 for the ORL dataset, 100 for the YaleB dataset.

**noise:** is used to set the noise you want to choose, you can choose to pass in "block_occlusion", "salt_and_pepper" or "no_noise"

**k:** is the number of component. In our report, we chose 10, 20, 30, and 40 as observation objects. You can also modify the k value to observe the prediction results

**size**: is used to randomly select a certain proportion of data for training. In the report, we chose 90%.

## 8.2 Team member contribution

| Task/Responsibility | Cheng Zeng | Yixuan Xu | Jiaqi Tian |
|---|---|---|---|
| **Report** | Noises<br><br>Preprocessing<br><br>Method of Robust NMF With L1 regularization<br><br>Instruction for code | Abstract<br><br>Introduction<br><br>Conclusion and future work<br><br>Method of L2NMF | Experiment<br><br>Method of L1NMF |

| Code | L1 NMF<br><br>Preprocessing<br><br>Block-occlusion noise<br><br>Code structure combing and integration | L21 NMF | L1NMF<br><br>Salt and pepper noise |
|---|---|---|---|